\documentclass{article} 
\usepackage{graphicx}
\usepackage{iclr2025_conference,times}

\usepackage{amsmath,amsfonts,bm}

\def\eqref#1{equation~\ref{#1}}

\def\1{\bm{1}}

\DeclareMathAlphabet{\mathsfit}{\encodingdefault}{\sfdefault}{m}{sl}
\SetMathAlphabet{\mathsfit}{bold}{\encodingdefault}{\sfdefault}{bx}{n}

\usepackage{hyperref}
\usepackage{url}

\title{“It Listens Better Than My Therapist”: Exploring Social Media Discourse on LLMs as Mental Health Tool}

\author{Anna-Carolina Haensch \\
LMU Munich; University of Maryland, College Park\\
Ludwigstrasse 33, Munich, Germany \\
\texttt{C.Haensch@lmu.de}
}

\iclrfinalcopy 
\begin{document}

\maketitle

\textbf{Abstract:} The emergence of generative AI chatbots such as ChatGPT has prompted growing public and academic interest in their role as informal mental health support tools. While early rule-based systems have been around since several years, large language models (LLMs) offer new capabilities in conversational fluency, empathy simulation, and availability. This study explores how users engage with LLMs as mental health tools by analyzing over 10,000 TikTok comments from videos referencing LLMs as mental health tools. Using a self-developed tiered coding schema and supervised classification models, we identify user experiences, attitudes, and recurring themes. Results show that nearly 20\% of comments reflect personal use, with these users expressing overwhelmingly positive attitudes. Commonly cited benefits include accessibility, emotional support, and perceived therapeutic value. However, concerns around privacy, generic responses, and the lack of professional oversight remain prominent. It Is important to note that the user feedback  does not indicate which therapeutic framework, if any, the LLM-generated output aligns with. While the findings underscore the growing relevance of AI in everyday practices, they also highlight the urgent need for clinical and ethical scrutiny in the use of AI for mental health support.

\textit{This study does not endorse or encourage the use of AI tools as substitutes for professional mental health support. The findings are presented for research purposes only, and any interpretation should take into account the limitations and potential risks of relying on AI in mental health contexts.}

\section{Introduction}

The rise of AI chatbots like ChatGPT has sparked interest in their potential role in mental health support. With demand for therapy often exceeding available resources, some view chatbots as a means to help bridge gaps in access \citep{Melo2024}. Increasingly, individuals are turning to ChatGPT to discuss personal concerns, including anxiety and relationship challenges, raising questions about how its support compares to traditional face-to-face therapy. To explore this phenomenon, we have collected several thousand comments from TikTok that discuss the use of LLMs as mental health tool.

Before the rise of LLMs, rule-based chatbots have already shown promise in addressing some of the limitations of digital mental health interventions (DMHIs). These systems simulate human conversation through predefined scripts and structured decision-making algorithms, such as decision trees, enabling interactive and dynamic mental health support \citep{Vaidyam2019,Lim2022}. Well-known rule-based chatbots like Woebot and Wysa have been found to alleviate depression symptoms \citep{Fitzpatrick2017,Inkster2018} and establish therapeutic alliances comparable to those formed with human therapists \citep{Beatty2022} (see also the systematic review by \citep{Farzan2024}). Moreover, qualitative studies suggest that users appreciate the human-like interactions and the sense of social support these chatbots provide \citep{Malik2022}. However, despite their potential, rule-based chatbots remain limited in fully realizing the promise of DMHIs. A meta-analysis by \citep{He2023} indicates that their therapeutic effects tend to be small and often do not persist over time.

Unlike rule-based chatbots, generative AI systems—such as OpenAI’s ChatGPT, Meta's Llama or Deepseek are trained on vast datasets, allowing them to generate and interpret language with remarkable fluency. These models are increasingly reaching or even surpassing human performance benchmarks in persuasive communication \citep{salvi2024conversationalpersuasivenesslargelanguage}, user engagement and responsiveness \citep{Yin2024}, and cognitive reframing to alleviate distress \citep{Li2024}. The rapid adoption of generative AI chatbots further highlights their potential impact, with ChatGPT reaching 100 million weekly active users within a year of its launch \citep{Malik2023} and an estimated half of the U.S. population having engaged with generative AI \citep{Salesforce2023}. 

The capabilities of generative AI may present a opportunity for digital mental health, with increasing media coverage and early research suggesting that these models may be more effective than rule-based chatbots in reducing psychological distress  in certain sitations \citep{Li2024}. However, their widespread adoption also introduces new challenges, including risks of harm, liability concerns, and trust-related issues such as AI hallucinations, lack of interpretability, and biases in training data \citep{Pandya2024}. Researchers are also exploring how design factors influence user comfort and disclosure. Interestingly, making a chatbot appear more human-like is not always a straightforward advantage. One 2023 study varied the profile image of an AI psychotherapy chatbot (from a company logo, to a robot avatar, to a human avatar, to a real human photo) and examined users’ willingness to self-disclose. The results showed that a more anthropomorphic (human-like) avatar did not automatically increase users’ self-disclosure \citep{Lee2023}. The clinical effectiveness of these models therefore remains a subject of ongoing investigation.

Despite growing public and academic interest, research in this area remains limited. Given the novelty of generative AI and the early stage of this field, further studies are essential to understand the implications, effectiveness, and risks associated with these technologies. In this context, social media data provides an exciting opportunity to complement ongoing small-scale interview studies and surveys \citep{Alanezi2024, Maples2024,Siddals2024}. While structured studies offer valuable insights into controlled settings, \textit{large-scale social media discussions allow us to observe how people organically describe AI chatbots in real-world situations}. By combining these approaches, we can develop a more comprehensive understanding of how generative AI is shaping mental health support and identify the gaps between expectations, experiences, and potential risks.

\section{LLMs as a Mental Health Tool Discussed by Mental Health Professionals}

AI chatbots offer several potential benefits as mental health support tools, particularly in terms of accessibility. Available 24/7 without the need for appointments, they provide immediate responses, making support more reachable for individuals facing long wait times, financial barriers, or limited access to professional help \citep{WHO2023Depression}. 

Another advantage is the perceived privacy and anonymity they offer. Some users may feel more comfortable discussing sensitive topics with an AI rather than a human, as there is no fear of judgment \citep{LUCAS201494}. This can encourage honest self-disclosure, particularly in cases where stigma might otherwise discourage open discussion, such as with addiction or suicidal thoughts. The absence of social evaluation can make AI chatbots a non-intimidating space for individuals seeking emotional support (for privacy concerns see below).

Although AI does not experience emotions, many chatbots are designed to simulate supportive communication. They often respond with validation and encouragement. Studies by mental health professionals have found their recommendations to be helpful, though often limited in complexity \citep{eshghie2023chatgpttherapistassistantsuitability}. 

 On the other hand, their limitations highlight the importance of human professionals in therapeutic settings. One of the most significant challenges is the lack of human connection.  The therapeutic alliance—an essential factor in successful treatment—relies on a therapist’s capacity to provide warmth, non-verbal cues, and personal engagement, aspects that AI cannot fully replicate \citep{horvath1991relation}. Unlike human therapists, AI often does not actively ask follow-up questions about an individual’s history, identity, or deeper psychological struggles, contributing to advice that often remains surface-level and generic \citep{Kimmel2023}. Most models rely heavily on cognitive behavioral therapy (CBT) techniques, which, while effective for many, do not encompass the full spectrum of therapeutic approaches \citep{Roklicer2025}. 
 
 Researchers have also directly compared AI-driven counseling with human therapy. For example, one trial had participants engage in “therapeutic” conversations with either a chatbot or a trained clinician. Initial findings suggest that while AI can deliver practical advice and engage users, it does not match the depth of human therapists in fostering trust and emotional connection (leading researchers to describe AI support as a helpful tool, but still “tentative” compared to real therapy) \citep{pham2022artificial}. A recent systematic review similarly found that various mental health chatbots can effectively reduce anxiety or depression symptoms, but many of these tools lack rigorous validation \citep{eshghie2023chatgpttherapistassistantsuitability}.

The risk of errors or harmful advice is another grave concern. LLM therapists do not guarantee accuracy and may produce misleading, overly simplistic, or inappropriate responses. While AI models are designed to avoid harmful recommendations, there have been instances where automated systems failed to recognize mental health crises or offered advice that was poorly suited to a person’s specific needs \citep{Pirnay2023}.  Researchers have highlighted concerns that AI could unintentionally reinforce systemic biases, leading to advice that is less inclusive or relevant for certain populations \citep{DAIR2023}.  Research also suggests that automated assessments may underestimate risk levels compared to human clinicians, potentially leading to inadequate support. AI’s content moderation filters may also prevent open discussion of high-risk topics, meaning users struggling with suicidal thoughts, trauma, or severe mental illness may receive generic warnings rather than meaningful guidance. Beyond accuracy, AI lacks clinical accountability. Unlike licensed therapists, AI operates without oversight from professional regulatory bodies and is not bound by confidentiality laws. If an AI fails to detect warning signs of a severe mental health condition, there is no formal mechanism to ensure user safety or hold anyone responsible for potential harm. 

Privacy and data security are additional concerns when using AI for mental health support. Unlike human therapists, who adhere to strict confidentiality regulations, AI systems process and store user conversations, raising questions about how sensitive information is handled. While some platforms allow users to disable chat history storage, many individuals may not be aware of these settings, increasing the risk of unintended data exposure.

\section{LLMs as a Mental Health Tool Discussed by Users and Patients}

Early studies provide mixed evidence about the effectiveness of LLMs as a mental health aid. A small pilot trial in a psychiatric inpatient setting found that patients who chatted with ChatGPT (3–6 guided sessions) reported improved quality of life scores versus a control group, along with high satisfaction with the AI sessions. In fact, participants’ satisfaction averaged 26.8 out of 30 on a Likert scale, indicating very positive feedback  \citep{Melo2024}.

Another empirical study focused on anxiety disorders had 399 outpatients use ChatGPT for 4 weeks and then surveyed them. The majority found the AI helpful – 91.2\% felt its responses were accurate, and many characterized the guidance as effective and coherent. Notably, this study reported that female participants tended to rate ChatGPT’s advice as more trustworthy and useful than male participants did. At the same time, users flagged significant concerns: about 67\% worried about privacy and 65\% had ethical reservations about using AI in this context. The authors concluded that ChatGPT shows promise as a complement to traditional therapy (enhancing access to care), but not a standalone replacement \citep{Alanzi2024}.

In short, current research indicates ChatGPT can provide meaningful support – sometimes improving users’ well-being – yet these studies are preliminary. All emphasize that human therapists remain irreplaceable for now, and that more research is needed on AI’s long-term efficacy and safety \citep{eshghie2023chatgpttherapistassistantsuitability}.

\section{Data and Methods Including Previous Research with TikTok Data}

\paragraph{TikTok as a Platform}

TikTok is a globally popular social media platform designed for sharing short-form videos. Originally launched in China in 2016 as Douyin, the platform expanded internationally after acquiring and rebranding the app musical.ly as TikTok in 2018. It currently boasts over a billion monthly active users worldwide \citep{statista_social_media_2023}. TikTok allows users to create and share videos ranging from 15 seconds to 10 minutes, providing various creative editing tools such as visual filters, musical soundtracks, special effects, and editing templates. One distinguishing feature is its powerful recommendation algorithm, which differentiates TikTok from other platforms by focusing less on personal network content and more on highly tailored recommendations based on user interactions and content preferences \citep{anderson2020getting}. The platform enjoys particular popularity among young audiences; in the United States, approximately 32.5\% of users are between 10–19 years old, and another 29.5\% are aged 20–29 \citep{clement2020ustiktok}.

\paragraph{Research Utilizing TikTok Data}

 A systematic review by \citet{mccashin2023using} highlights diverse research topics studied through TikTok data, including COVID-19 awareness, dermatology, eating disorders, cancer awareness, neurological tics, radiology education, sexual health, genetic information dissemination, and broader public health communication. While most research originates from the United States, significant contributions also emerge from China, Ireland, Australia, and Canada.

Recent data underscores TikTok’s prominence among American teenagers aged 18–19, with approximately 67\% reporting regular platform usage \citep{statista_tiktok_2021}. Usage among college-aged adults (20–29 years) is slightly lower but still significant, estimated at around 56\% \citep{statista_tiktok_2021}. Nevertheless, researchers must acknowledge potential sampling limitations, as the TikTok user population might not fully represent broader demographic subgroups or the general population.

\paragraph{Sampling and Sample Characteristics}

On March 3, 2025, we collected data using a newly created and previously unused TikTok account. For each of the search terms — “chatgpt therapist,” “chatbot therapist,” “AI therapist,” and “LLM therapist” — we retrieved the IDs of the top-listed videos. As we progressed through the results, we observed a growing number of videos unrelated to the topic of "LLMs as mental health support." These videos were often included due to hashtag farming or because they addressed only one of the two dimensions — either large language models or mental health, but not both. Data collection for each term was therefore stopped after encountering 10 such unrelated videos. We then used the TikTok Research API to collect the comments and user engagement statistics for the identified 85 videos on March 6, 2025. 

We observed a wide variation in the popularity of the videos included in our dataset. Among the 83 videos with comments, some had minimal engagement, while others reached viral status. For instance, the number of shares ranged from 0 to over 578,000, with a median of 34 but a mean of 10,668, indicating that a few highly shared videos skewed the average.

Similarly, view counts ranged from just 396 to more than 10 million, with a median of 11,724 and a mean of 422,200, again reflecting a long-tailed distribution. Like counts varied widely as well, from 13 to 1.5 million, with a median of 445 and a mean of 48,902. 

These statistics illustrate that while most videos attracted moderate engagement, a small number achieved exceptionally high visibility and interaction, highlighting the heterogeneous nature of TikTok content performance even within a focused topical search.

\begin{table}[ht]
\centering
\caption{Descriptive Statistics for Videos with Comments}
\label{tab:with_comments}
\begin{tabular}{lrrrrrr}
\hline
Metric         & Mean   & Median & Min   & Max       & Q1    & Q3 \\
\hline
View Count      & 422200 & 11724   &   396  & 10900000  & 4078   & 120446 \\
Like Count      & 48902  &   445   &    13  & 1500000   &  124   & 7570 \\
Share Count     & 10668  &    34   &     0  & 578800    &   10   & 826 \\
Comment Count   &   305  &    24   &     1  & 6046      &    8   & 144 \\
\hline
\end{tabular}
\end{table}

\section{Developing a Coding Schema to Capture User Perspectives and Experiences}

Understanding user experiences with AI chatbots in mental health therapy through comments on social media platforms such as TikTok requires a robust coding schema that systematically captures both positive and negative perspectives. This process involves defining key categories based on qualitative and quantitative analysis of user feedback. Our approach is informed by prior research, particularly the studies by \citet{Chaudhry2024}, which identified themes in user comments on an AI-driven mental health conversationalist application; and \citet{Pandya2024}, which critically discussed problematic characteristics of LLMs as therapists. However, based on an initial screening of 200 social media comments, we added new categories with topics that users mentioned. 

\section{Tiered Classification Schema}

We developed a structured coding schema to systematically classify user comments into clearly defined categories. This schema facilitates capturing nuanced user perspectives and their experiences with ChatGPT as a tool for mental health support. The classification unfolds in three hierarchical stages, moving progressively from general identification of personal use towards thematic characterizations of positive and negative user perceptions. This structured approach ensures comprehensive coverage of user attitudes and facilitates meaningful qualitative and quantitative analysis of the data.

\subsection{Stage 1: Own Experience (Binary)}
This first level determines whether the user has personally used ChatGPT as a mental health support tool:
\begin{itemize}
    \item \textbf{Yes (1):} The commenter explicitly mentions having used ChatGPT as a therapist or for emotional support.
    \item \textbf{No (0):} No personal experience mentioned.
\end{itemize}

\subsection{Stage 2: Opinion on Using ChatGPT as a Therapist}
The second level determines the attitude towards using ChatGPT for therapy/mental health support/emotional support tool.
\begin{itemize}
    \item \textbf{In Favor (+1):} Positive attitude
    \item \textbf{Against (-1):} Negative attitude
    \item \textbf{Neutral (0):} Neither positive nor negative or mixed attitudes.
\item \textbf{Not Applicable (-99):} No opinion expressed. This includes random or off-topic comments, as well as those that only mention another account (e.g., tagging a user) or contain only emojis without substantive content.

\end{itemize}

\subsection{Stage 3: Characteristics of User Perspectives}
Once experience and stance are established, comments that voice an opinion on LLMs as therapy/mental health support/emotional support tool are further categorized based on specific characteristics:\\

\textbf{(Perceived as) Positive Characteristics:}
    \begin{itemize}
        \item \textbf{Always Available:} Users highlight AI tools’ constant, 24/7 availability. Some appreciate their patience and reliability, often contrasting them with the unpredictability of human interactions.

        \item \textbf{Low-Cost or Free:} Users view AI tools as an accessible, cost-free, or low-cost alternative to professional mental health services.

        \item \textbf{Emotional Outlet:} Users describe AI as a safe, non-judgmental space for venting, self-reflection, and comfort—sometimes referring to the tool as a friend or companion.

        \item \textbf{AI as Therapist:} Users report receiving therapeutic guidance and mental health advice from AI, at times explicitly referring to it as a therapist or therapist substitute.

        \item \textbf{Memory \& Continuity:} Users note AI’s ability to recall past conversations and maintain continuity, which enhances the feeling of ongoing support.
    \end{itemize}
    
\textbf{(Perceived as) Negative Characteristics:}
    \begin{itemize}
        \item \textbf{Paywall \& Access Limits:} Users express frustration with usage caps, limited free access, or subscription fees that restrict continued use.

        \item \textbf{Superficial Understanding:} Users question whether AI truly understands them, noting that LLMs generate likely responses rather than demonstrating genuine comprehension.

        \item \textbf{Generic or Robotic Replies:} Users find responses to be repetitive, impersonal, or emotionally flat, lacking the nuance of human interaction.

        \item \textbf{Over-Affirming Behavior:} Users criticize AI for being too agreeable, avoiding pushback or critical engagement when needed.

        \item \textbf{Not a Real Therapist:} Users emphasize that AI lacks the lived experience, training, and emotional depth of human therapists.

        \item \textbf{Referral to Professionals:} Users note that AI often suggests seeking human therapists but express frustration when real-life access is limited or unavailable.

        \item \textbf{Misleading or Unhelpful Advice:} Users report instances where advice was inaccurate, overly simplistic, or poorly suited to their needs, sometimes causing emotional discomfort.

        \item \textbf{Privacy Concerns:} Users worry about how their personal and sensitive data might be stored, used, or accessed by third parties.

        \item \textbf{No Accountability:} Users question who is responsible if the AI provides harmful or inappropriate guidance.

        \item \textbf{Overdependence:} Users express concern about becoming too emotionally reliant on AI for support or decision-making.

        \item \textbf{Therapist Job Loss:} Users raise ethical concerns about AI potentially replacing human mental health professionals.

        \item \textbf{Environmental Impact:} Users mention AI’s energy and water consumption, highlighting concerns about environmental sustainability.
    \end{itemize}

For manual annotation of users’ own experiences, user attitudes, and topical relevance,  900 of the most popular comments based on the number of replies received were selected. This purposive sampling approach was chosen over random sampling because many of the less popular comments consisted primarily of low-content interactions, such as user mentions or emoji-only responses. These comments typically did not contribute substantive opinions or experiences regarding using AI in mental health contexts. By focusing on high-engagement comments, the sample was more likely to include meaningful discourse and explicit stances and facilitating the fine-tuning process of the coding LLM.

Three supervised fine-tuning runs on the gpt-4o-2024-08-06 base model were conducted to classify the comments in regard to users’ own experiences, user attitudes, and topical relevance. All models were trained for 3 epochs with a batch size of 1. For the experience classifier, the model was trained on 800 manually labeled examples and validated on 100 instances. The loss curve showed rapid early convergence, with stable token-level accuracy across both training and validation sets. The attitude classifier was trained on 800 examples (approx. 360,000 tokens) and validated on 100. The model demonstrated consistent performance, with minimal fluctuations in loss and stable token accuracy throughout. The topic labeling model was fine-tuned on 399 examples (approx. 200,000 tokens). While training loss showed greater variability, token-level accuracy remained consistently high.  Importantly, all topic-labeled comments were manually verified by the author. In contrast, for the experience and attitude models, only subsets of the data were reviewed in detail. Consequently, any potential misestimation in the prevalence of topics is more likely to result in a modest undercount than in an overstatement.

\begin{figure}
    \centering
    \includegraphics[width=0.7\linewidth, trim=0cm 30cm 0cm 6cm, clip]{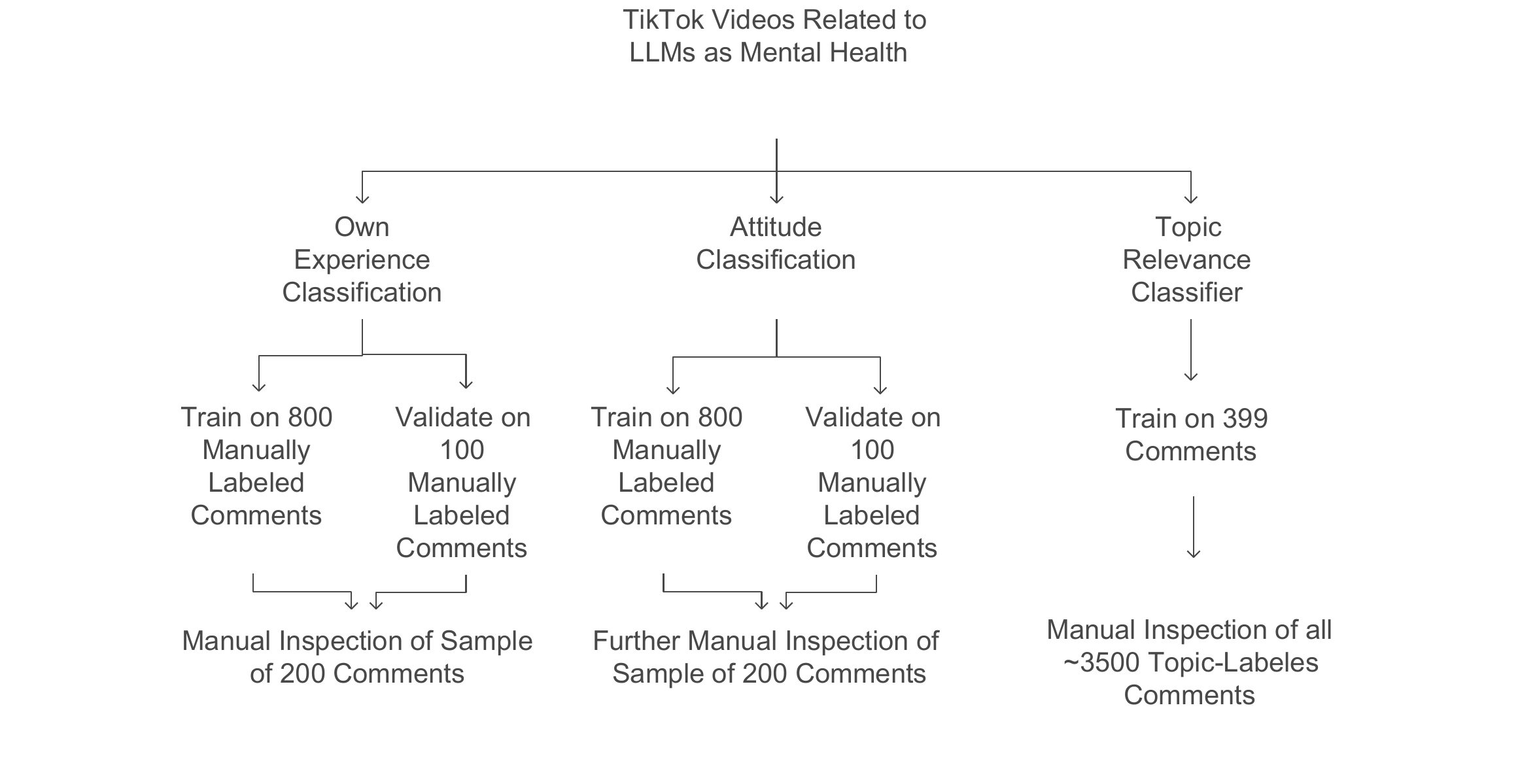}
    \caption{Classification pipeline}
    \label{fig:enter-label}
\end{figure}

\section{Results}

\subsection{Usage of AI as a mental health tool and general attitudes}

 Based on our classification pipeline, 19.65\%  of all responses (N=9,923) were identified as reflecting a personal experience with AI, while 80.35\% did not mention such an experience. Among the classifiable responses for the attitude towards AI as a mental health tool, 20.18\% were classified as positive, 7.74\% as negative, and 6.58\% as neutral. These proportions highlight a generally positive tilt among those with explicit opinions, though negative and neutral stances are also meaningfully represented. It is important to note that the substantial amount of unclassifiable content; i.e., posts without an identifiable attitude toward AI as a mental health tool (mentions of other AI use cases were excluded) or without discernible experience—often consisted of superficial engagement signals such as @mentions, emojis, or generic expressions. These lacked sufficient semantic depth to allow for reliable interpretation.
 
  When we look deeper at the correlation between experiences and attitude, commenters who have used an AI tool for emotional support overwhelmingly tend to express a positive stance, whereas those with no direct experience are more divided, with relatively more skepticism. In fact, among the 1950 comments from people with firsthand experience with LLMs as a mental health support tool, a large majority voiced favorable opinions. By contrast, the comments from non-users include a higher proportion of neutral or negative views. The ratio of positive to negative comments and experience vs. non-experience should not be interpreted as a reflection of attitudes in the general population. Instead, it simply shows that the topic is receiving attention and that there is a substantial group of users who actively engage with and make use of LLMs as a mental health tool. The presence of both supportive and critical voices suggests that the technology is relevant enough to provoke discussion, but it does not allow for conclusions about exact user statistics in the general population.
For example, the Pew Research Center survey found that 79\% of U.S. adults say they would not want to use an AI chatbot for mental health support, underscoring how general attitudes can be quite negative in the abstract. 
Nevertheless, the user experiences reflected in these almost 2000 comments with firsthand experience with LLMs as mental health support tool deserve closer examination — especially given the current lack of large-scale studies on how people actually engage with LLMs for emotional support in real-world settings. The broader discussion on TikTok is also worth analyzing to understand what kinds of information and narratives are most visible to young people encountering this topic on the platform.

\subsection{Recurring themes and topics in the comments}

\begin{figure}
    \centering
    \includegraphics[width=1\linewidth, trim=0cm 15cm 0cm 0cm, clip]{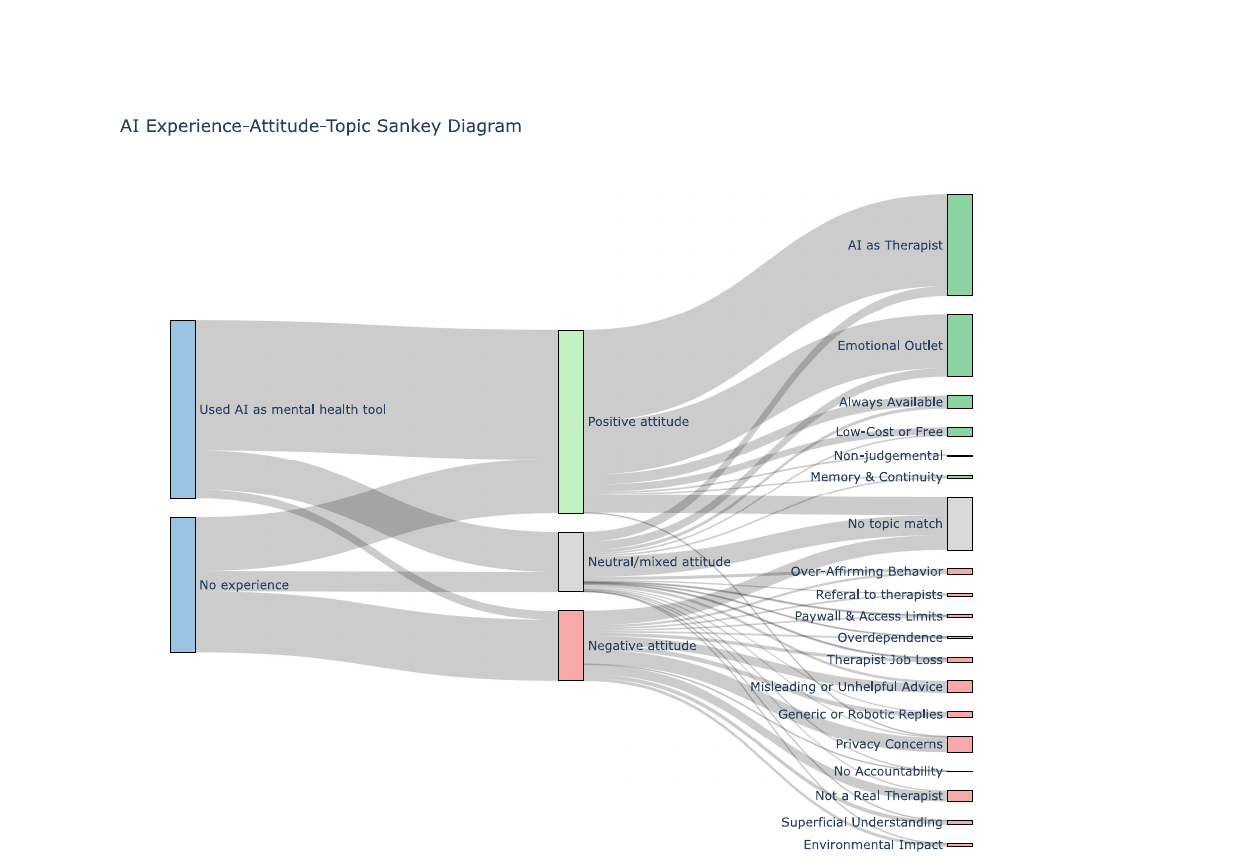}
    \caption{Sankey diagram of experiences, attitudes and topics mentioned. The comments that were not directly related to LLMs as mental health tool (code:-99) were filtered out here.}
    \label{fig:sankey}
\end{figure}

Positive attitudes centered on a few recurring supportive themes (see Figure \ref{fig:sankey}). Notably, two themes – AI as Therapist (1104 comments) and Emotional Outlet (678 comments) – stand out as the most frequently mentioned benefits. Together these account for roughly 42\% of all 4189 coded comments, indicating that the primary value users see in AI mental health tools is the ability to receive advice and feel heard or comforted. Many experienced users described getting useful suggestions, coping strategies, or perspective from the AI (AI as Therapist), as well as a sense of companionship or empathy (Emotional Outlet). These findings resonate with qualitative reports of users finding an emotional sanctuary in AI \cite{Siddals2024}.

Some commenters explicitly noted the non-judgmental support nature of AI (~20 mentions) and the benefit of AI remembering past conversations (~30 mentions), though these appeared less frequently.  Similarly, users acknowledge that an AI counselor can be available anytime, anywhere, without waiting for appointments, and often at little to no expense – a significant plus for those who cannot afford traditional therapy or need support outside office hours. The affordability factor (and by extension, accessibility) is clearly on some users’ minds. In summary, the discourse around positive experiences emphasizes AI as a helpful advisor and compassionate companion, with practical advantages like constant availability and low cost mentioned but not at the forefront.

Although negative stances were a minority, the data reveal several distinct concerns that fuel skepticism or criticism of AI for mental health. The most prominent worry by far is Privacy, with around 180 comments voicing concerns about data confidentiality, security, or the sensitive nature of sharing one’s personal thoughts with an AI.  Many people who have not used such tools preemptively cite privacy as a barrier to trust. This finding aligns with survey research showing that people are more concerned about the privacy of their mental health information when using an AI platform than when speaking to a human counselor.

Another widely echoed caution is that AI should not be seen as a full replacement for human therapists. The topic on AI being not a real human garnered more than 100 mentions, making it the second-largest negative theme. Comments in this vein often acknowledge some utility in AI tools but stress that AI lacks the human qualities necessary for serious therapy, for example, empathy, genuine understanding, or the ability to handle complex crises. Many argue that while an AI chatbot might be a helpful supplement, it cannot replicate the nuanced empathy and accountability of a trained human professional.

The quality of AI’s responses was another area of criticism, particularly among those who had tried the tools. Misleading or unhelpful advice was cited in 128 comments, indicating instances where the advice or information given by the AI was viewed as ineffective, generic, or off-target. Similarly, around 60 comments complained about generic or robotic replies, and the same number of comments noted that the AI was over-affirming their behavior. The users felt the AI just gave easy affirmations or generic positivity without truly engaging or challenging when needed. Some 30 comments worry about individuals becoming too dependent on AI support.

A few participants looked at AI mental health tools through a broader societal lens. A handful of comments raised the issue of who is responsible if the AI gives bad advice or if something goes wrong. This suggests that ethical or legal accountability is not top-of-mind for most lay commenters, perhaps because more immediate experience-based issues like privacy and usefulness dominate. Likewise,  potential job loss for therapists was a secondary concern (around 50 comments): Finally, around 40  comments mentioned the environmental impact, alluding to the environmental costs of running AI systems (energy usage/carbon footprint).

\section{Summary and Discussion}

\subsection{Summary}

The rapid emergence of generative AI chatbots, particularly large language models (LLMs) like ChatGPT, has sparked renewed interest in their role as mental health support tools. While previous rule-based systems such as Woebot and Wysa demonstrated modest but limited success, LLMs show promise in offering more natural, persuasive, and responsive interactions. Despite these advances, key concerns remain about their therapeutic depth, accuracy, and ethical implications. Our study explores this evolving discourse by analyzing over 10,000 TikTok comments referencing LLMs as therapy substitutes. We developed a tiered classification schema to capture users’ self-reported experiences, attitudes, and thematic concerns.

Among the 1950 users who reported personal use, the majority expressed positive sentiments, especially regarding AI’s accessibility, emotional support, and perceived therapeutic value. Yet, the analysis also uncovered notable concerns, especially regarding data privacy, lack of genuine understanding, and the superficial, over-affirming or generic nature of AI responses.  The scale and tone of TikTok discussions highlight both the perceived utility and persistent skepticism surrounding AI in mental health contexts. Social media data, while limited in representativeness, provides valuable insight into real-world engagement and emerging norms. Our findings underscore the need for further research to evaluate the effectiveness, risks, and regulatory implications of LLM-based mental health support in everyday use.

\subsection{Limitations}

While the analysis offers important insights into public perceptions of LLMs as mental health tools, several limitations must be acknowledged. First, the dataset is drawn from TikTok, a platform with a young and demographically skewed user base, limiting the generalizability of findings. Commenters are self-selecting, and high-engagement content may overrepresent extreme or humorous viewpoints. Second, although we applied a structured coding schema and fine-tuned models for classification, not all comments could be reliably interpreted due to low semantic content, and annotation bias remains a possibility. Importantly, expressed attitudes do not necessarily reflect actual mental health outcomes. Positive sentiment may reflect momentary satisfaction rather than long-term benefit, while critical views may stem from unfamiliarity rather than informed evaluation. Furthermore, while social media analysis offers scale and immediacy, it cannot substitute for clinical trials or in-depth interviews that assess therapeutic effectiveness. Finally, ethical concerns—such as AI accountability, misinformation, and the commercialization of mental health support—are only partly addressed in user discourse and require more rigorous, interdisciplinary investigation.

\section{Acknowledgment}

This research was conducted using data accessed through the TikTok Research API. We acknowledge TikTok's support in providing access to public data for academic research purposes. The findings, interpretations, and conclusions expressed in this work are those of the authors and do not necessarily reflect the views of TikTok or its affiliates.

\section{Disclaimer}

This study does not endorse or encourage the use of AI tools as substitutes for professional mental health support. The findings are presented for research purposes only, and any interpretation should take into account the limitations and potential risks of relying on AI in mental health contexts.

\bibliography{iclr2025_conference}

\begin{thebibliography}{35}
\providecommand{\natexlab}[1]{#1}
\providecommand{\url}[1]{\texttt{#1}}
\expandafter\ifx\csname urlstyle\endcsname\relax
  \providecommand{\doi}[1]{doi: #1}\else
  \providecommand{\doi}{doi: \begingroup \urlstyle{rm}\Url}\fi

\bibitem[Alanezi(2024)]{Alanezi2024}
Fahad Alanezi.
\newblock Assessing the effectiveness of chatgpt in delivering mental health support: A qualitative study.
\newblock \emph{Journal of Multidisciplinary Healthcare}, Volume 17:\penalty0 461–471, January 2024.
\newblock ISSN 1178-2390.
\newblock \doi{10.2147/jmdh.s447368}.
\newblock URL \url{http://dx.doi.org/10.2147/JMDH.S447368}.

\bibitem[Alanzi et~al.(2024)Alanzi, Alharthi, Alrumman, Abanmi, Jumah, Alansari, Alharthi, Alibrahim, Algethami, Aburass, Alshahrani, Alzahrani, Alotaibi, Magadmi, and Almasodi]{Alanzi2024}
Turki~M Alanzi, Abdulaziz Alharthi, Sarah Alrumman, Sobhia Abanmi, Ammar Jumah, Hatun Alansari, Taif Alharthi, Abdulrahman Alibrahim, Abdullah Algethami, Mishaal Aburass, Abdullah~Mohammed Alshahrani, Shahad Alzahrani, Batool Alotaibi, Talah Magadmi, and Mohammed~Saeed Almasodi.
\newblock Chatgpt as a psychotherapist for anxiety disorders: An empirical study with anxiety patients.
\newblock \emph{Nutrition and Health}, October 2024.
\newblock ISSN 2047-945X.
\newblock \doi{10.1177/02601060241281906}.
\newblock URL \url{http://dx.doi.org/10.1177/02601060241281906}.

\bibitem[Anderson(2020)]{anderson2020getting}
K.~E. Anderson.
\newblock Getting acquainted with social networks and apps: It is time to talk about tiktok.
\newblock \emph{Library Hi Tech News}, 37\penalty0 (4):\penalty0 7--12, 2020.
\newblock \doi{10.1108/lhtn-01-2020-0001}.

\bibitem[Beatty et~al.(2022)Beatty, Malik, Meheli, and Sinha]{Beatty2022}
Clare Beatty, Tanya Malik, Saha Meheli, and Chaitali Sinha.
\newblock Evaluating the therapeutic alliance with a free-text cbt conversational agent (wysa): A mixed-methods study.
\newblock \emph{Frontiers in Digital Health}, 4, April 2022.
\newblock ISSN 2673-253X.
\newblock \doi{10.3389/fdgth.2022.847991}.
\newblock URL \url{http://dx.doi.org/10.3389/fdgth.2022.847991}.

\bibitem[Chaudhry \& Debi(2024)Chaudhry and Debi]{Chaudhry2024}
Beenish~Moalla Chaudhry and Happy~Rani Debi.
\newblock User perceptions and experiences of an ai-driven conversational agent for mental health support.
\newblock \emph{mHealth}, 10:\penalty0 22–22, July 2024.
\newblock ISSN 2306-9740.
\newblock \doi{10.21037/mhealth-23-55}.
\newblock URL \url{http://dx.doi.org/10.21037/mhealth-23-55}.

\bibitem[Clement(2020)]{clement2020ustiktok}
J.~Clement.
\newblock U.s. tiktok users by age 2020.
\newblock \url{https://www.statista.com/statistics/1095186/tiktok-ususers-age}, 2020.
\newblock Accessed on February 15, 2023.

\bibitem[Dixon(2023)]{statista_social_media_2023}
S.~Dixon.
\newblock Most popular social networks worldwide as of january 2023, ranked by number of monthly active users.
\newblock \url{https://www.statista.com/statistics/272014/global-social-networks-ranked-by-number-of-users/}, 2023.
\newblock [Online; accessed March 27, 2023].

\bibitem[Eshghie \& Eshghie(2023)Eshghie and Eshghie]{eshghie2023chatgpttherapistassistantsuitability}
Mahshid Eshghie and Mojtaba Eshghie.
\newblock Chatgpt as a therapist assistant: A suitability study, 2023.
\newblock URL \url{https://arxiv.org/abs/2304.09873}.

\bibitem[Farzan et~al.(2024)Farzan, Ebrahimi, Pourali, and Sabeti]{Farzan2024}
Maryam Farzan, Hamid Ebrahimi, Maryam Pourali, and Fatemeh Sabeti.
\newblock Artificial intelligence-powered cognitive behavioral therapy chatbots, a systematic review.
\newblock \emph{Iranian Journal of Psychiatry}, December 2024.
\newblock ISSN 2008-2215.
\newblock \doi{10.18502/ijps.v20i1.17395}.
\newblock URL \url{http://dx.doi.org/10.18502/ijps.v20i1.17395}.

\bibitem[Fitzpatrick et~al.(2017)Fitzpatrick, Darcy, and Vierhile]{Fitzpatrick2017}
Kathleen~Kara Fitzpatrick, Alison Darcy, and Molly Vierhile.
\newblock Delivering cognitive behavior therapy to young adults with symptoms of depression and anxiety using a fully automated conversational agent (woebot): A randomized controlled trial.
\newblock \emph{JMIR Mental Health}, 4\penalty0 (2):\penalty0 e19, June 2017.
\newblock ISSN 2368-7959.
\newblock \doi{10.2196/mental.7785}.
\newblock URL \url{http://dx.doi.org/10.2196/mental.7785}.

\bibitem[He et~al.(2023)He, Yang, Qian, Li, Su, Zhang, and Hou]{He2023}
Yuhao He, Li~Yang, Chunlian Qian, Tong Li, Zhengyuan Su, Qiang Zhang, and Xiangqing Hou.
\newblock Conversational agent interventions for mental health problems: Systematic review and meta-analysis of randomized controlled trials.
\newblock \emph{Journal of Medical Internet Research}, 25:\penalty0 e43862, April 2023.
\newblock ISSN 1438-8871.
\newblock \doi{10.2196/43862}.
\newblock URL \url{http://dx.doi.org/10.2196/43862}.

\bibitem[Horvath \& Symonds(1991)Horvath and Symonds]{horvath1991relation}
Adam~O Horvath and B~Dianne Symonds.
\newblock Relation between working alliance and outcome in psychotherapy: A meta-analysis.
\newblock \emph{Journal of counseling psychology}, 38\penalty0 (2):\penalty0 139, 1991.

\bibitem[Inkster et~al.(2018)Inkster, Sarda, and Subramanian]{Inkster2018}
Becky Inkster, Shubhankar Sarda, and Vinod Subramanian.
\newblock An empathy-driven, conversational artificial intelligence agent (wysa) for digital mental well-being: Real-world data evaluation mixed-methods study.
\newblock \emph{JMIR mHealth and uHealth}, 6\penalty0 (11):\penalty0 e12106, November 2018.
\newblock ISSN 2291-5222.
\newblock \doi{10.2196/12106}.
\newblock URL \url{http://dx.doi.org/10.2196/12106}.

\bibitem[Institute(2023)]{DAIR2023}
Distributed AI~Research Institute.
\newblock Dair statement on march 2023 open letter calling for ai development pause, March 2023.
\newblock URL \url{https://www.dair-institute.org/blog/letter-statement-March2023/}.
\newblock Accessed: 2025-03-04.

\bibitem[Kimmel(2023)]{Kimmel2023}
Daniel Kimmel.
\newblock {ChatGPT Therapy Is Good, But It Misses What Makes Us Human}.
\newblock \emph{Columbia University Department of Psychiatry}, 2023.
\newblock URL \url{https://www.columbiapsychiatry.org/news/chatgpt-therapy-is-good-but-it-misses-what-makes-us-human}.

\bibitem[Lee \& Lee(2023)Lee and Lee]{Lee2023}
Jieon Lee and Daeho Lee.
\newblock User perception and self-disclosure towards an ai psychotherapy chatbot according to the anthropomorphism of its profile picture.
\newblock \emph{Telematics and Informatics}, 85:\penalty0 102052, November 2023.
\newblock ISSN 0736-5853.
\newblock \doi{10.1016/j.tele.2023.102052}.
\newblock URL \url{http://dx.doi.org/10.1016/j.tele.2023.102052}.

\bibitem[Li et~al.(2024)Li, Herderich, and Goldenberg]{Li2024}
Joanna~Z. Li, Alina Herderich, and Amit Goldenberg.
\newblock Skill but not effort drive gpt overperformance over humans in cognitive reframing of negative scenarios.
\newblock \emph{Center for Open Science}, April 2024.
\newblock \doi{10.31234/osf.io/fzvd8}.
\newblock URL \url{http://dx.doi.org/10.31234/osf.io/fzvd8}.

\bibitem[Lim et~al.(2022)Lim, Shiau, Cheng, and Lau]{Lim2022}
Shi~Min Lim, Chyi Wey~Claudine Shiau, Ling~Jie Cheng, and Ying Lau.
\newblock Chatbot-delivered psychotherapy for adults with depressive and anxiety symptoms: A systematic review and meta-regression.
\newblock \emph{Behavior Therapy}, 53\penalty0 (2):\penalty0 334–347, March 2022.
\newblock ISSN 0005-7894.
\newblock \doi{10.1016/j.beth.2021.09.007}.
\newblock URL \url{http://dx.doi.org/10.1016/j.beth.2021.09.007}.

\bibitem[Lucas et~al.(2014)Lucas, Gratch, King, and Morency]{LUCAS201494}
Gale~M. Lucas, Jonathan Gratch, Aisha King, and Louis-Philippe Morency.
\newblock It’s only a computer: Virtual humans increase willingness to disclose.
\newblock \emph{Computers in Human Behavior}, 37:\penalty0 94--100, 2014.
\newblock ISSN 0747-5632.
\newblock \doi{https://doi.org/10.1016/j.chb.2014.04.043}.
\newblock URL \url{https://www.sciencedirect.com/science/article/pii/S0747563214002647}.

\bibitem[Malik(2023)]{Malik2023}
A.~Malik.
\newblock Openai’s chatgpt now has 100 million weekly active users, 2023.
\newblock URL \url{https://techcrunch.com/2023/11/06/openais-chatgpt-now-has-100-million-weekly-active-users/}.
\newblock TechCrunch AI, accessed on March 5, 2025.

\bibitem[Malik et~al.(2022)Malik, Ambrose, and Sinha]{Malik2022}
Tanya Malik, Adrian~Jacques Ambrose, and Chaitali Sinha.
\newblock Evaluating user feedback for an artificial intelligence–enabled, cognitive behavioral therapy–based mental health app (wysa): Qualitative thematic analysis.
\newblock \emph{JMIR Human Factors}, 9\penalty0 (2):\penalty0 e35668, April 2022.
\newblock ISSN 2292-9495.
\newblock \doi{10.2196/35668}.
\newblock URL \url{http://dx.doi.org/10.2196/35668}.

\bibitem[Maples et~al.(2024)Maples, Cerit, Vishwanath, and Pea]{Maples2024}
Bethanie Maples, Merve Cerit, Aditya Vishwanath, and Roy Pea.
\newblock Loneliness and suicide mitigation for students using gpt3-enabled chatbots.
\newblock \emph{npj Mental Health Research}, 3\penalty0 (1), January 2024.
\newblock ISSN 2731-4251.
\newblock \doi{10.1038/s44184-023-00047-6}.
\newblock URL \url{http://dx.doi.org/10.1038/s44184-023-00047-6}.

\bibitem[McCashin \& Murphy(2023)McCashin and Murphy]{mccashin2023using}
D.~McCashin and CM~Murphy.
\newblock Using tiktok for public and youth mental health--a systematic review and content analysis.
\newblock \emph{Clinical Child Psychology and Psychiatry}, 28\penalty0 (1):\penalty0 279--306, 2023.
\newblock \doi{10.1177/13591045221106608}.

\bibitem[Melo et~al.(2024)Melo, Silva, and Lopes]{Melo2024}
Antonio Melo, In\^es Silva, and Joana Lopes.
\newblock Chatgpt: A pilot study on a promising tool for mental health support in psychiatric inpatient care.
\newblock \emph{International Journal of Psychiatric Trainees}, 2\penalty0 (2), February 2024.
\newblock ISSN 3005-3870.
\newblock \doi{10.55922/001c.92367}.
\newblock URL \url{http://dx.doi.org/10.55922/001c.92367}.

\bibitem[Organization(2023)]{WHO2023Depression}
World~Health Organization.
\newblock Depressive disorder, 2023.
\newblock URL \url{https://www.who.int/news-room/fact-sheets/detail/depression}.
\newblock Accessed: 2025-03-04.

\bibitem[Pandya et~al.(2024)Pandya, Lodha, and Ganatra]{Pandya2024}
Apurvakumar Pandya, Pragya Lodha, and Amit Ganatra.
\newblock Is chatgpt ready to change mental healthcare? challenges and considerations: a reality-check.
\newblock \emph{Frontiers in Human Dynamics}, 5, January 2024.
\newblock ISSN 2673-2726.
\newblock \doi{10.3389/fhumd.2023.1289255}.
\newblock URL \url{http://dx.doi.org/10.3389/fhumd.2023.1289255}.

\bibitem[Pham et~al.(2022)Pham, Nabizadeh, and Selek]{pham2022artificial}
Kay~T Pham, Amir Nabizadeh, and Salih Selek.
\newblock Artificial intelligence and chatbots in psychiatry.
\newblock \emph{Psychiatric Quarterly}, 93\penalty0 (1):\penalty0 249--253, 2022.

\bibitem[Pirnay(2023)]{Pirnay2023}
Emma Pirnay.
\newblock {We Spoke to People Who Started Using ChatGPT As Their Therapist}.
\newblock \emph{VICE}, 2023.
\newblock URL \url{https://www.vice.com/en/article/we-spoke-to-people-who-started-using-chatgpt-as-their-therapist/}.

\bibitem[Roklicer(2025)]{Roklicer2025}
Laura Roklicer.
\newblock {ChatGPT in Psychotherapy: 3 Case Studies}.
\newblock \emph{Psychology Today}, 2025.
\newblock URL \url{https://www.psychologytoday.com/intl/blog/lucid-story/202501/chatgpt-in-psychotherapy-3-case-studies}.

\bibitem[Salesforce(2023)]{Salesforce2023}
Salesforce.
\newblock Top generative ai statistics for 2024, 2023.
\newblock URL \url{https://www.salesforce.com/news/stories/generative-ai-statistics/}.
\newblock Salesforce News \& Insights, accessed on March 5, 2025.

\bibitem[Salvi et~al.(2024)Salvi, Ribeiro, Gallotti, and West]{salvi2024conversationalpersuasivenesslargelanguage}
Francesco Salvi, Manoel~Horta Ribeiro, Riccardo Gallotti, and Robert West.
\newblock On the conversational persuasiveness of large language models: A randomized controlled trial, 2024.
\newblock URL \url{https://arxiv.org/abs/2403.14380}.

\bibitem[Siddals et~al.(2024)Siddals, Torous, and Coxon]{Siddals2024}
Steven Siddals, John Torous, and Astrid Coxon.
\newblock “it happened to be the perfect thing”: experiences of generative ai chatbots for mental health.
\newblock \emph{npj Mental Health Research}, 3\penalty0 (1), October 2024.
\newblock ISSN 2731-4251.
\newblock \doi{10.1038/s44184-024-00097-4}.
\newblock URL \url{http://dx.doi.org/10.1038/s44184-024-00097-4}.

\bibitem[{Statista}(2023)]{statista_tiktok_2021}
{Statista}.
\newblock Tiktok: Distribution of {U.S.} users 2022, by age group.
\newblock \url{https://www.statista.com/statistics/1095186/tiktok-us-users-age/}, 2023.
\newblock [Online; accessed February 20, 2023].

\bibitem[Vaidyam et~al.(2019)Vaidyam, Wisniewski, Halamka, Kashavan, and Torous]{Vaidyam2019}
Aditya~Nrusimha Vaidyam, Hannah Wisniewski, John~David Halamka, Matcheri~S. Kashavan, and John~Blake Torous.
\newblock Chatbots and conversational agents in mental health: A review of the psychiatric landscape.
\newblock \emph{The Canadian Journal of Psychiatry}, 64\penalty0 (7):\penalty0 456–464, March 2019.
\newblock ISSN 1497-0015.
\newblock \doi{10.1177/0706743719828977}.
\newblock URL \url{http://dx.doi.org/10.1177/0706743719828977}.

\bibitem[Yin et~al.(2024)Yin, Jia, and Wakslak]{Yin2024}
Yidan Yin, Nan Jia, and Cheryl~J. Wakslak.
\newblock Ai can help people feel heard, but an ai label diminishes this impact.
\newblock \emph{Proceedings of the National Academy of Sciences}, 121\penalty0 (14), March 2024.
\newblock ISSN 1091-6490.
\newblock \doi{10.1073/pnas.2319112121}.
\newblock URL \url{http://dx.doi.org/10.1073/pnas.2319112121}.

\end{thebibliography}
\bibliographystyle{iclr2025_conference}

\newpage

\appendix
\section{Appendix}

\clearpage
\begin{table}
\centering
\caption{Experience × Attitude Crosstab }
\label{tab:experience_attitude}
\begin{tabular}{lrr}
\hline
Own experience y &  No experience &  Used AI as mental health tool \\
Attitude &                &                                \\
\hline
Negative attitude          &            674 &                             94 \\
Neutral/mixed attitude     &            219 &                            434 \\
Positive attitude          &            585 &                           1417 \\
\hline
\end{tabular}
\end{table}

\begin{table}
\centering
\caption{Own Experience × Topic Crosstab}
\label{tab:experience_topic}
\begin{tabular}{lrr}
\hline
Own experience y &  No experience &  Used AI as mental health tool \\
Topic       &                &                                \\
\hline
AI as Therapist                &            267 &                            837 \\
Always Available               &             48 &                             96 \\
Emotional Outlet               &            147 &                            531 \\
Environmental Impact           &             33 &                              2 \\
Generic or Robotic Replies     &             50 &                             13 \\
Low-Cost or Free               &             38 &                             57 \\
Memory \& Continuity            &             19 &                             15 \\
Misleading or Unhelpful Advice &             75 &                             53 \\
No Accountability              &              4 &                              1 \\
No topic match                 &            350 &                            227 \\
Non-judgemental                &              6 &                             12 \\
Not a Real Therapist           &            109 &                             12 \\
Over-Affirming Behavior        &             33 &                             26 \\
Overdependence                 &             21 &                              6 \\
Paywall \& Access Limits        &             19 &                              9 \\
Privacy Concerns               &            160 &                             18 \\
Referal to therapists          &              9 &                             25 \\
Superficial Understanding      &             42 &                              4 \\
Therapist Job Loss             &             48 &                              1 \\
\hline
\end{tabular}
\end{table}

\begin{table}
\centering
\caption{Attitude × Topic Crosstab}
\label{tab:attitude_topic}
\begin{tabular}{lrrr}
\hline
Attitude &  Negative attitude &  Neutral/mixed attitude &  Positive attitude \\
Topic       &                    &                         &                    \\
\hline
AI as Therapist                &                  0 &                     108 &                996 \\
Always Available               &                  0 &                      35 &                109 \\
Emotional Outlet               &                  0 &                      93 &                585 \\
Environmental Impact           &                 33 &                       2 &                  0 \\
Generic or Robotic Replies     &                 49 &                      14 &                  0 \\
Low-Cost or Free               &                  0 &                      18 &                 77 \\
Memory \& Continuity            &                  0 &                      18 &                 16 \\
Misleading or Unhelpful Advice &                 98 &                      30 &                  0 \\
No Accountability              &                  4 &                       1 &                  0 \\
No topic match                 &                161 &                     219 &                197 \\
Non-judgemental                &                  0 &                       0 &                 18 \\
Not a Real Therapist           &                105 &                      16 &                  0 \\
Over-Affirming Behavior        &                 25 &                      34 &                  0 \\
Overdependence                 &                 19 &                       8 &                  0 \\
Paywall \& Access Limits        &                 17 &                      11 &                  0 \\
Privacy Concerns               &                159 &                      15 &                  4 \\
Referal to therapists          &                 19 &                      15 &                  0 \\
Superficial Understanding      &                 43 &                       3 &                  0 \\
Therapist Job Loss             &                 36 &                      13 &                  0 \\
\hline
\end{tabular}
\end{table}

\end{document}